\documentclass[journal,twoside,web]{ieeecolor}
\usepackage{generic}
\usepackage{cite}
\usepackage{amsmath,amssymb,amsfonts}
\usepackage{algorithmic}
\usepackage{graphicx}
\usepackage{multicol}
\usepackage{multirow}
\usepackage{textcomp}
\usepackage{amsmath,amsfonts}
\usepackage{algorithm}
\usepackage{array}
\usepackage{mathtools}
\usepackage{eqnarray,amssymb}
\usepackage{booktabs,siunitx} 
\usepackage{bm}
\usepackage{stfloats}
\usepackage{url}
\usepackage{verbatim}

\usepackage{graphicx}
\usepackage{epstopdf}

\def\BibTeX{{\rm B\kern-.05em{\sc i\kern-.025em b}\kern-.08em
    T\kern-.1667em\lower.7ex\hbox{E}\kern-.125emX}}
\begin{document}
\title{A Predictive Model Based on Transformer with Statistical Feature Embedding in Manufacturing Sensor Dataset}

\author{Gyeong Taek Lee and Oh-Ran Kwon
\thanks{Gyeong Taek Lee is an Assistant professor in the Department of Mechanical, Smart, and Industrial Engineering, Gachon University, Seongnam, 13120, Republic of Korea (e-mail: leegt@gachon.ac.kr).}
\thanks{Oh-Ran Kwon is a Postdoctoral researcher in the Department of Data Sciences and Operations at the University of Southern California Marshall School of Business, Los Angeles, CA 90089, USA (e-mail: ohrankwo@marshall.usc.edu).} 
}

\maketitle

\begin{abstract}

In the manufacturing process, various sensor data is collected from the equipment. Engineers can use this collected data to build predictive models, enabling them to manage the process and improve production yield. However, due to the rapid development of process technology, collecting a sufficient amount of data has become challenging, making it difficult to create powe
rful models. In this study, we propose a novel predictive model based on the Transformer architecture by developing statistical feature embedding and window positional encoding. Statistical features are an effective representation of sensor data in the manufacturing domain. We introduce statistical feature embedding to allow the Transformer to effectively learn both time- and sensor-related information. We create statistical feature embedding by applying several different statistical-pooling layers individually to the sensor data and combining the outputs. Additionally, window positional encoding can capture precise time information from the proposed feature embedding. We assess the performance of the proposed model in two different problems: fault detection and virtual metrology. Experimental results demonstrate that the novel predictive model outperforms baseline models. This can be attributed to efficient parameter usage—an advantage for sensor data, which is inherently limited in sample size. Our result supports the applicability of our model across various manufacturing industries.

\end{abstract}


\section{Introduction}
\label{sec:introduction}
\IEEEPARstart{I}{n} industry 4.0, a stable smart manufacturing technology system is essential to ensure the competitiveness of manufacturing companies. This requires a high-level quality monitoring system or powerful predictive model. Stable systems aid in developing intelligent manufacturing through quality monitoring and predictive models. However, the rapid advancement of manufacturing technology has led to a notable increase in the complexity of manufacturing equipment and process recipes. This complexity has been further addressed by the emergence of a new generation of information technologies dealing with collected data from the equipment \cite{yan2023review}. In general, from the equipment or production line, a large amount of sensor data is collected. Therefore, there has been a large effort in building high-quality and stable monitoring systems or predictive models.

In manufacturing processes, there are several types of predictive models, including fault detection (FD) models, virtual metrology (VM) models, and yield prediction models. A FD model is a critical element in building a smart manufacturing system. Such a system is indispensable for accurately identifying and diagnosing faults within production equipment. As a result, interest in improving intelligence in FD models, diagnosis, and prediction is increasing. This increased emphasis on intelligent FD has led to applications in various domains, including battery manufacturing \cite{hu2020advanced,din2023automated}, semiconductor production \cite{lee2017convolutional,kim2019fault,fan2020data,hsu2021multiple}, automotive assembly \cite{li2009sensor,wang2021automatic}, panel manufacturing \cite{yang2009sensor,jin2012bayesian,popescu2020real} and numerous other production lines \cite{guo2021fault,peng2022fault}. In general, FD systems detect the abnormal status of production based on the status of sensor data from the equipment.

%For example, in the semiconductor manufacturing process, if gas pressure is unstable or shows a different trend compared to a normal production, the FD model can identify the production as being in an abnormal state.

To manage production quality effectively, a VM model is essential, particularly in semiconductor manufacturing where each lot comprises 10 to 25 wafers, with physical metrology typically conducted on only 1 or 2 wafers for measuring thickness, critical dimensions, and electrical resistance. These measurements are critical for ensuring wafer quality, necessitating the use of VM models developed from sensor data \cite{kang2009virtual,kang2011virtual,hwang2013robust,gentner2021dbam,clain2021virtual,maggipinto2019deepvm,chien2022decision}. Additionally, we can forecast the volume of production that can be delivered to the customer by predicting the yield \cite{wu2010fuzzy,yuan2011yield,jun2020quality,stich2020yield,lee2022attention}. 

Time-series data encompassing various values, such as humidity, gas, and pressure, is collected from the sensors at specific intervals from the equipment. Predictive models dealing with sensor data in manufacturing processes face significant challenges. As machine learning methods have evolved, various forms of these predictive models have emerged. Conventionally, many previous studies have extracted statistical features from time-series sensor data using a window, and they have utilized these extracted features in machine learning models such as linear regression, support vector machines, random forests (RF), and gradient boosting \cite{kang2009virtual,kang2011virtual,hwang2013robust,lee2022sequential,lee2023abnormal}. This approach has shown good performance, but it cannot learn graphical features from the sensor data and could cause a high-dimensionality problem. 

With the development of deep learning, various approaches have been emerging. The most basic approach to processing such sensor data is to view data with an x-axis sensor and a y-axis time as a 2D image and use it as an input of convolutional neural networks (CNN) that performs very well for image processing \cite{lee2017convolutional,kim2019fault,maggipinto2019deepvm,hsu2021multiple}. In addition, long short-term memory (LSTM), which has been widely used in sequential data such as time-series data and natural language processing (NLP), has been employed in building predictive models for manufacturing processes \cite{xiang2021fault,jalayer2021fault,belagoune2021deep,yuan2022sia}. These predictive models excel over the conventional approach because they can learn graphical features from time-series data. However, to achieve good performance, a deep learning model requires a large amount of data \cite{munir2018deepant,krishna2019deep,he2020model}. When the number of observations is small, it is highly likely that the conventional approach is superior to the deep learning model \cite{lee2023abnormal}. However, in the fields, it is difficult to obtain large amounts of data due to natural equipment wear or changing process recipes. Therefore, it is necessary to develop a predictive model that can capture graphical information from the time-series sensor data when the number of data is small.

Meanwhile, in the field of NLP, the Transformer has been proposed as an architecture based on attention mechanisms \cite{vaswani2017attention}. The Transformer is a neural network architecture that leverages self-attention and parallel processing to excel in sequence-to-sequence tasks. The attention mechanism allows models to focus on specific parts of the input sequence while making predictions, enabling the model to capture long-range dependencies and relationships in the data more effectively. The innovative design of the Transformer has had a profound impact on the field of deep learning and has been widely adopted in various NLP tasks. Furthermore, in recent years, various variants of the Transformer have been developed in diverse fields, such as tabular data \cite{huang2020tabtransformer, gorishniy2021revisiting,badaro2022transformers}, time-series prediction \cite{zhou2021informer,tang2021probabilistic,zerveas2021transformer}, and generative models \cite{lin2018st,jiang2021transgan,zhang2022styleswin}.

\begin{figure}[!t]
\centering
\includegraphics[scale=0.48]{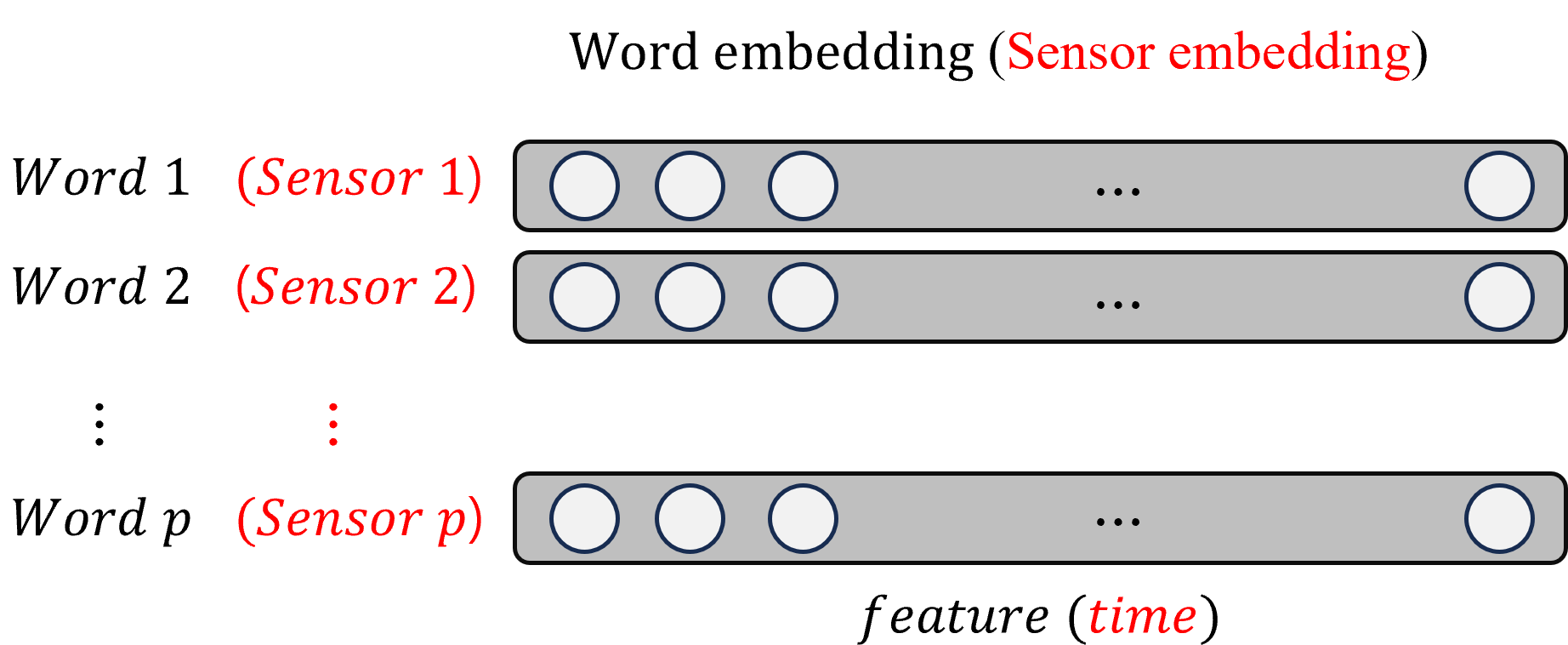}
\caption{Input embedding of sensor data.}
\label{fig1}
\vskip -0.2in
\end{figure}

The trend of utilizing the Transformer in various domains has now extended to time-series sensor data. One fundamental approach is to treat sensor variables as words \cite{zhang2022deep}, as illustrated in Fig \ref{fig1}. The model cannot accurately incorporate sequential information into the model since the sensor variable is not measured sequentially. When the sensor input is transposed, the Transformer can capture the sequential information \cite{liu2020novel,peng2022fault}. However, this model cannot effectively utilize sensor information. This is because, in the original Transformer, the column indicates the dimension of embedding to represent a word. In NLP, the most important information is in `words'. Similarly, in the sensor domain, the most critical information is in the `sensors'. Therefore, both sequential information of sensor should be simultaneously integrated into a model. One paper proposed a method to extract statistical features from sensors and utilize the LSTM layer within a Transformer \cite{liu2020novel}. However, this method also struggles to properly learn sensor information, and it requires too large a number of parameters. 

In this study, we propose a novel predictive model based on Transformer for the manufacturing sensor domain. To learn both sensor information and the sequential data from each sensor, we introduce a new method to organize the input embedding named statistical feature-based embeddings. It consists of the x-axis representing sensor information at a given time window and the y-axis representing statistical features for each sensor at that time. This sensor embedding is instrumental in capturing the sequential information of sensor variables. Additionally, we introduce a new positional encoding method tailored to this sensor embedding. The main contributions of this study are summarized as follows: 

\begin{itemize}
\item A novel Transformer model for time-series sensor data in the manufacturing domain is proposed. This model can effectively learn graphical features from time-series sensor data, even when the amount of available data is small. This point is supported by the results of our experiments, where our proposed model demonstrates impressive performance compared to existing methods on two real-world datasets.
\item The process of creating statistical feature embedding can be understood as applying several different statistical pooling layers individually to the sensor data and then combining the outputs. Thanks to statistical feature-based embedding, our proposed Transformer requires a substantially lower number of parameters compared to existing methods.
\item A new positional encoding method, designed for the sensor input embedding proposed in this work, can enhance the model's performance compared to conventional positional encoding. 
\item We conduct extensive comparisons of methodologies. We compare the proposed Transformer with various methods from other existing approaches, including conventional machine learning methods, deep learning-based methods (CNN and LSTM), and Transformer-based methods. 
\end{itemize}

The remaining parts of this paper are structured as follows: Section 2 reviews the Transformer in detail. Section 3 introduces our proposed model including sensor embedding and positional encoding. Subsequently, Section 4 presents the experiments and results. Additionally, Section 5 discusses the merits of our proposed model and potential future research. Finally, Section 6 concludes this paper.

\section{Background}
The Transformer \cite{vaswani2017attention} is one of the most well-known architectures in NLP. In this section, we review five main components of the Transformer. That are positional encoding, multi-headed attention, residual connection, normalization, and fully connected feed-forward network. These components serve as inspiration for, or have been adopted in, our proposed model architecture.

\begin{figure}[!t]
\centering
\includegraphics[scale=0.31]{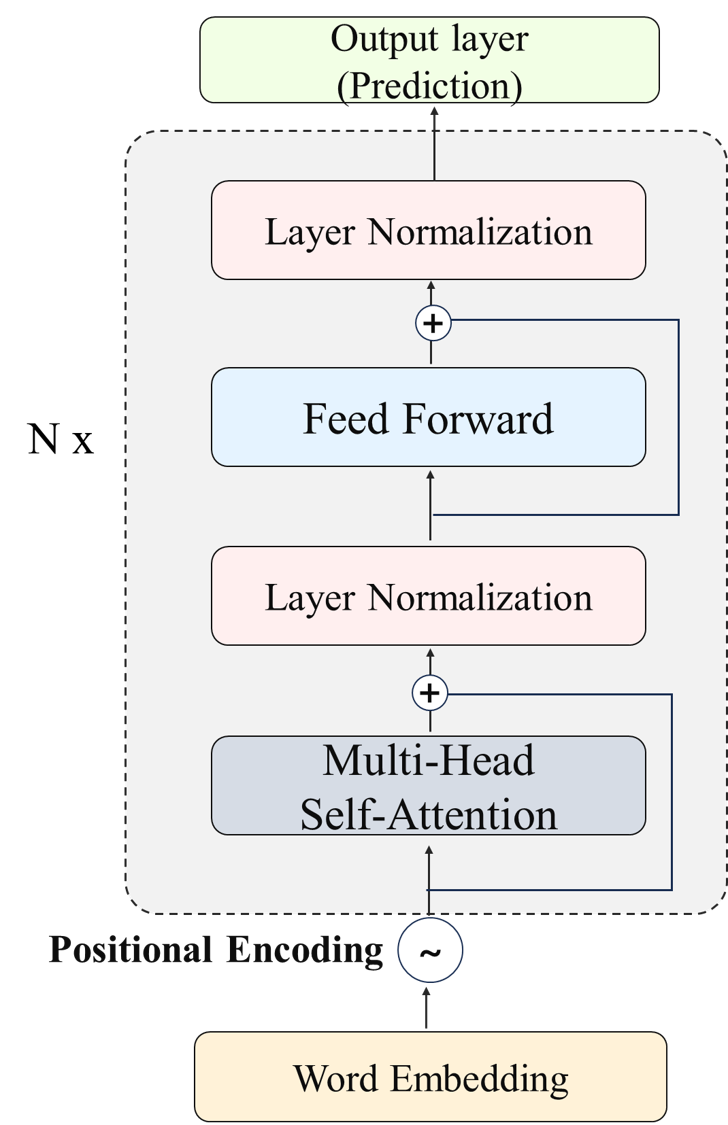}
\caption{The Transformer architecture.}
\label{fig:transformer}
\vskip -0.2in
\end{figure}

Fig \ref{fig:transformer} illustrates the simplified transformer architecture to facilitate an exposition of these components and their roles within the Transformer. At first, the positional encodings are added to the input embeddings. Then, the encoded input embeddings sequentially passes multi-head attention, normalization, feedforward network, and normalization layers. 
%the Transformer applies a multi-headed attention operation over the encoded input embeddings followed by a feedforward network layer.
A residual connection is used around the multi-headed attention and the feed forward network layers. The block composing from a multi-headed attention to the feedforward network is called a transformer block, which is the gray block in Fig \ref{fig:transformer}. Several transformer blocks can be concatenated and form an encoder-decoder structure. 

In the following subsections, we review the positional encoding in Subsection \ref{subsec:posit-encod} and the other four components inside the transformer block in Subsection \ref{subsec:transform-block}.

\subsection{Positional Encoding}\label{subsec:posit-encod}

The Transformer was originally applied to sequential data like a sentence in NLP. The input tokens such as words in a sentence are converted to vectors of dimension $d$ through the embedding layer. Let $\mathbf X_{emb} \in \mathbb R^{n \times d}$ be the embedded input where $n$ is the number of tokens and $d$ is the number of embedding features. Although input tokens are in sequential order, the transformer block does not take the sequential relationship into consideration unlike recurrent neural networks. Therefore, the position encoding matrix $\mathbf P \in \mathbb R^{n \times d}$ is introduced which purely contains the positional information:
\begin{equation*}
    \begin{aligned}
        & \mathbf P_{(pos, 2i)} = \sin (pos/10000^{2i/d}) \\
        \text{and } & \mathbf P_{(pos, 2i+1)} = \cos (pos/10000^{2i/d}),
    \end{aligned}
\end{equation*}
where $\mathbf M_{(i,j)}$ denotes the $(i,j)$-th element of the matrix $\mathbf M$. The positional information is injected in $\mathbf X_{\text{emb}}$ by adding $\mathbf P$:
\begin{equation}\label{eq:X}
    \mathbf X = \mathbf X_\text{emb} + \mathbf P.
\end{equation}
The positional encoding motivates the window positional encoding in our proposed model. 

\subsection{Transformer Block}\label{subsec:transform-block}

The transformer block is composed of four layers along with two residual connections. In the following four subsubsections, multi-headed attention, residual connection, normalization, and feedforward network are introduced, respectively. Then, the overall process of the transformer block from the input $\mathbf X$ to the output is summarized in Subsubsection \ref{subsub:overall-transformer} by putting four components together. 

\subsubsection{Multi-headed Attention}\label{subsub:attention}
Three types of multi-headed attention are used in the Transformer: multi-headed self-attention, masked multi-headed self-attention, and multi-headed encoder-decoder attention. The multi-headed self-attention is used in the encoder of the Transformer and the others in the decoder. Our review focuses on the multi-headed self-attention as it has been directly utilized in our proposed model architecture.

The multi-headed self-attention layer can be seen as applying a function $\text{LayerMHSA: } \mathbb R^{n\times d} \rightarrow \mathbb R^{n \times d}$ on the input, say $\mathbf M$, and produces $\text{LayerMHSA}(\mathbf M)$ where
\begin{equation*}
    \begin{aligned}
        & \mathbf Q^{(i)} = \mathbf M \mathbf W_i^Q, ~ \mathbf K^{(i)} = \mathbf M \mathbf W_i^K, \mathbf V^{(i)} = \mathbf M \mathbf W_i^V, \\
        & \mathbf H_i = \text{softmax}\left( \frac{\mathbf Q^{(i)} \left( \mathbf K^{(i)} \right)^T }{ \sqrt{d} } \right) \mathbf V^{(i)}, ~ \text{for } i=1,\ldots,h, \\ %\text{Attention} (\mathbf Q^{(h)}, \mathbf K^{(h)}, \mathbf V^{(h)}) \\
        \text{and } & \text{LayerMHSA}(\mathbf M) = \text{Concat}(\mathbf H_1,\ldots,\mathbf H_h) \mathbf W^O.
    \end{aligned}
\end{equation*}
The dimension of the parameter matrices are $\mathbf W_i^Q \in \mathbb R^{d \times k}$, $\mathbf W_i^K \in \mathbb R^{d \times k}$, $\mathbf W_i^V \in \mathbb R^{d \times k}$, and $\mathbf W^O\in\mathbb R^{h\cdot k\times d}$. The hyper parameters $h$ and $k$ are set to have $h\cdot k = d$. %The parameter $\mathbf w$ contains the entries of $\mathbf W_i^Q$, $\mathbf W_i^K$, $\mathbf W_i^V$, and $\mathbf W^O$ for all $i =1,\ldots,h$.
%In the literature, $\mathbf Q$, $\mathbf K$, and $\mathbf V$ are referred to as query, key, and value, respectively. 
The head $\mathbf H_i$ is computed by a scaled dot-product attention operation on $\mathbf Q^{(i)}$, $\mathbf K^{(i)}$, and $\mathbf V^{(i)}$.

\subsubsection{Residual Connection}
The residual connection \cite{he2016deep} is done by adding the input and the output of a layer. In Fig \ref{fig:transformer}, the residual connection is illustrated as an arrow connecting the input and the output of a layer with a plus sign in the end of the arrow. Let us say $\mathbf M$ is the input and $\text{Layer}(\mathbf M)$ is the output of a layer. Then, the residual connection is expressed as follows:
$$
    \mathbf M + \text{Layer}(\mathbf M).
$$

\subsubsection{Normalization}
%Next, the multi-headed attention is followed by a residual connection \cite{he2016deep} and a normalization \cite{ba2016layer}:
The normalization layer \cite{ba2016layer} is a simple method to reduce the training time of various neural network models. It can be expressed as a function $\text{LayerNorm: }\mathbf R^{n\times d} \rightarrow \mathbf R^{n\times d}$, where 
%\begin{equation*}
%    \begin{aligned}
%    \mathbf Z = \text{LayerNorm}(\mathbf X + \mathbf Z').
%    \end{aligned}
%\end{equation*}
the $i$-th row of $\text{LayerNorm}(\mathbf M)$ is defined as 
$$
    [\text{LayerNorm}(\mathbf M)]_{(i,\cdot)} = \bm\gamma^T \circ \frac{\mathbf M_{(i,\cdot)} - \bm \mu_i}{\bm \sigma_i} + \bm \beta^T.
$$
Here, $\mathbf M_{(i,\cdot)}$ denotes the $i$-th row of $\mathbf M$, $\bm\gamma,\bm \beta \in \mathbb R^{d}$ are parameters, 
$$
    \bm \mu_i = \textstyle\sum_j \mathbf M_{(i,j)} / d, ~~ \bm \sigma_i = \sqrt{\textstyle\sum_j ( \mathbf M_{(i,j)} - \bm\mu_i )^2/d },
$$
and $\circ$ is the element-wise multiplication between two vectors.

\subsubsection{Position-wise Feed-Forward Networks}\label{subsub:ffnn}
%In the second layer, the output from the first layer $\mathbf Z$ passes the feedfoward network followed by the residual connection and normailization as follows. 
The position-wise feedforward network layer applies the same fully-connected feedforward network to each position. %It consists of two linear transformations and a ReLu activation. 
It can be defined as a function $\text{LayerFFN: } \mathbb R^{n\times d}\rightarrow \mathbb R^{n\times d}$ %on $\mathbf Z$ and produces $g_{\mathbf w_2}(\mathbf Z) = \mathbf U'$ 
where the $i$-th row of $\text{LayerFFN}(\mathbf M)$ is defined as 
$$
    [\text{LayerFFN}(\mathbf M)]_{(i,\cdot )} = \max(0,  \mathbf M_{(i,\cdot)} \mathbf W_1 + \mathbf b_1^T ) \mathbf W_2 + \mathbf b_2^T.
    %\text{FN}(\mathbf Z) = \max(0,  \mathbf Z_{(i,\cdot) \mathbf W_1} + \mathbf b_1^T ) \mathbf W_2 + \mathbf b_2^T
$$
The dimensions of parameters are $\mathbf W_1 \in \mathbb R^{d \times m}$, $\mathbf W_2 \in \mathbb R^{m\times d}$, $\mathbf b_1 \in \mathbb R^{m}$, and $\mathbf b_2 \in \mathbb R^{d}$.

\subsubsection{Overall Mechanism}\label{subsub:overall-transformer}
The transformer block takes $\mathbf X$, defined in \eqref{eq:X}, as the input. Let us explore the output of the transformer block using the four components introduced above. The input $\mathbf X \in \mathbb R^{n\times d}$ passes the multi-head self attention followed by the residual connection and the normalization and produces $\mathbf Z\in\mathbb R^{n\times d}$:
$$
    \mathbf Z = \text{LayerNorm} ( \mathbf X + \text{LayerMHSA} (\mathbf X) ) .
$$
Then the feedforward network is applied on $\mathbf Z$, again followed by the residual connection and the normalization:
$$
    \mathbf U = \text{LayerNorm} ( \mathbf Z + \text{LayerFFN} (\mathbf Z) ),
$$
which is the output of the transformer block. As the output $\mathbf U \in \mathbb R^{n\times d}$ has the same dimension as the input $\mathbf X$, several transformer blocks can be stacked, each with its own set of parameters.

\section{Proposed Model Architecture}
\begin{figure}[!t]
\centering
\includegraphics[scale=0.58]{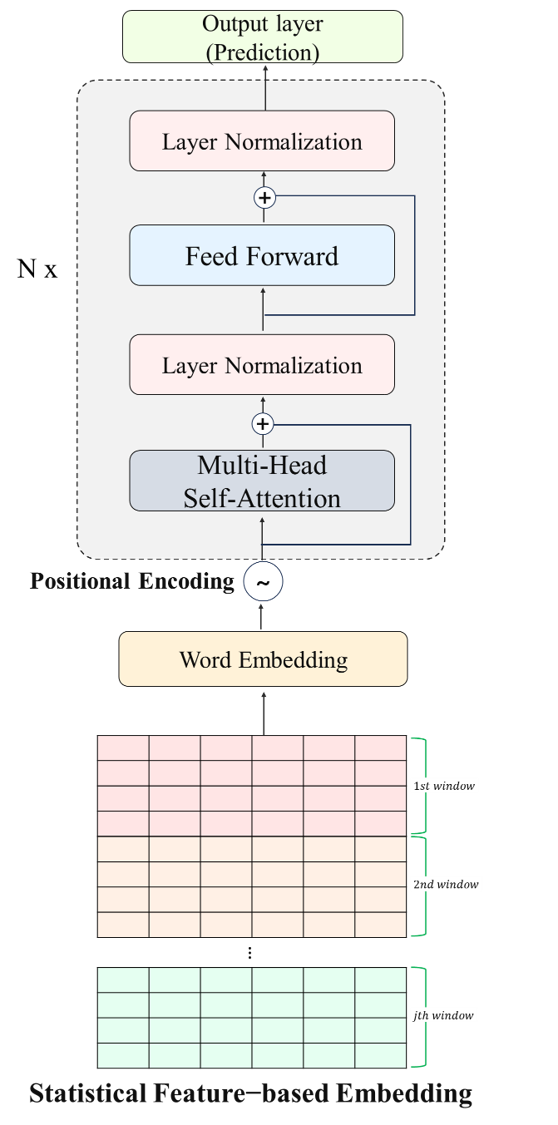}
\caption{The architecture of the proposed model. }
\label{fig:proposed}
\vskip -0.2in
\end{figure}

In this section, we introduce the proposed model architecture for manufacturer  time-series sensor data, which is described in Fig \ref{fig:proposed}. Each component is explained in the following subsections.

%In this study, we propose a novel Transformer for the manufacturing sensor domain. To learn both sensor information and the sequential data from each sensor, we organize the input embedding. It consists of the x-axis representing sensor information at a given time window and the y-axis representing statistical features for each sensor at that time. This sensor embedding is instrumental in capturing the sequential information of sensor variables. Additionally, we introduce a new positional encoding method tailored to this sensor embedding. The main contributions of this study are summarized as follows:

\subsection{Statistical Feature Embedding}\label{subsec:stat-embed}
%In the Transformer in the context of NLP, the sequential data like words in a sentence are converted to two dimensional input matrix $\mathbf X_{emb} \in \mathbb R^{n\times d}$. Each column is $d$ number of features of a token and rows are in a sequential order of tokens. 
To effectively train both sensor and sequential information in the sensor data through the Transformer in the context of manufacturer time-series sensor data, we develop the statistical feature embedding.

\begin{figure}[!t]
\centering
\includegraphics[scale=0.41]{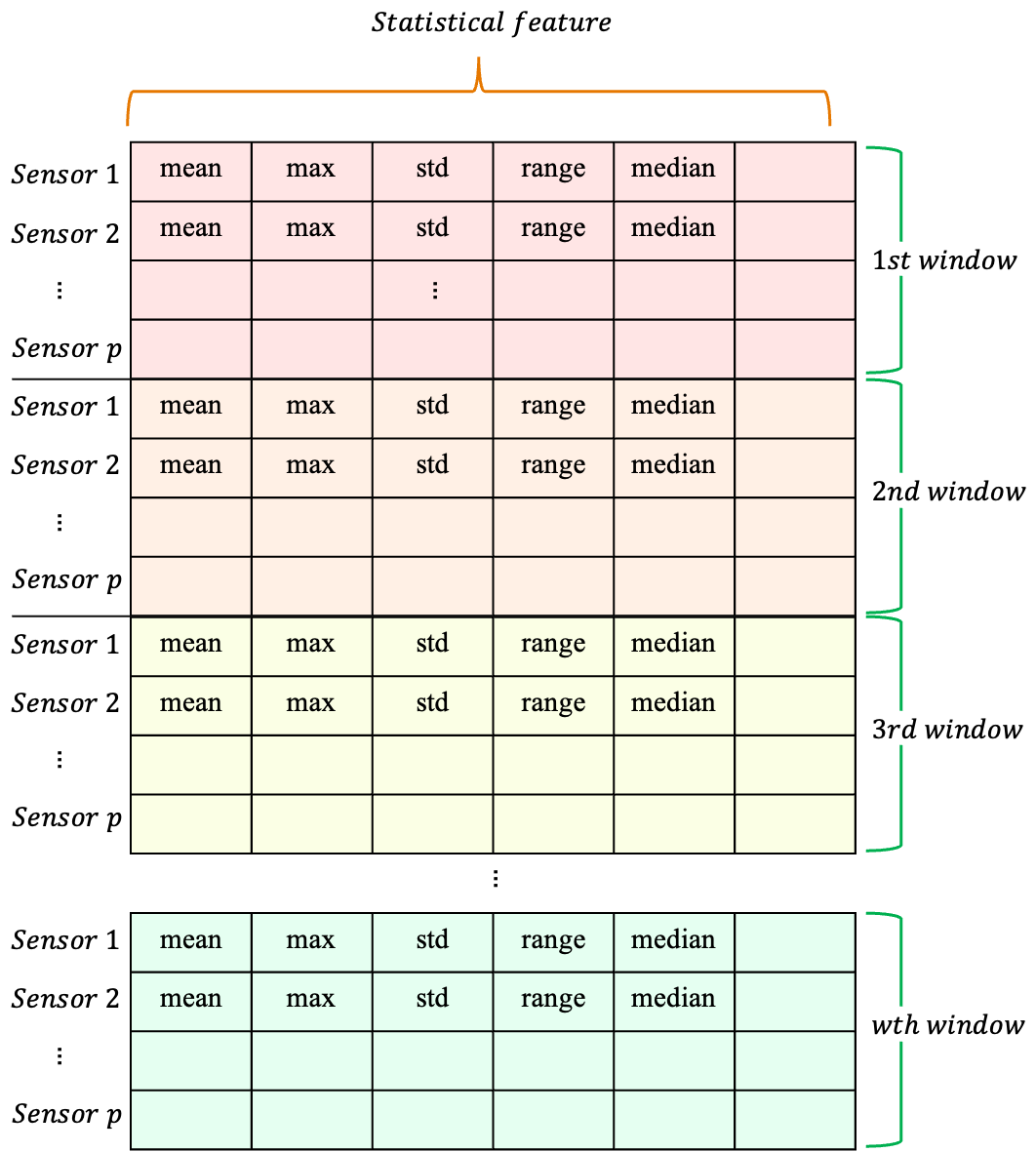}
\caption{Statistical feature-based embedding of sensor data.}
\label{fig:embedding}
\vskip -0.2in
\end{figure}

In manufacturing processes, $p$ number of sensors are measured from time $1$ to time $t$ for each sample. 
%Let $s_{ij}$ be a $i$-th sensor data point at time $j$ where $i=1,\ldots, p$ and $j=1,\ldots,t$. 
We partition the time into $w$ number of time window blocks, say $I_1,\ldots,I_w$ such that $\cup_i I_i = \{1,\ldots,t\}$. Each time window contains a similar number of consecutive time points and they do not overlap, i.e., $n_I = |I_i| = |I_j|$ and $I_i \cap I_j = \emptyset$ for any $i\neq j$. 
In each time window, we compute statistical features for each sensor. The examples of statistical features are mean, max, and standard deviation. Let us say that $d$ number of statistical features are considered and $s_{ijk}$ is the $k$-th statistical feature of the $i$-th sensor in the $j$-th time window. %If the first statistical feature is a mean, $s_{ij1}$ is the mean of $i$-th sensor data points in $j$-th time interval. %$= \sum_{b \in I_j} s_{ij} / |I_j| $, the average of $\{ s_{ib} : b \in I_j \}$.
We organize the calculated statistical features in the following manner.
\begin{equation}\label{eq:x-stat-emb}
    \mathbf X_{\text{stat-emb}} \in \mathbb R^{n \times d},
\end{equation}
where $[\mathbf X_{\text{stat-embed}}]_{(i+p(j-1),k)} = s_{ijk}$, which is illustrated in Fig \ref{fig:embedding}. %is the $k$-th statistical feature of $i$-th sensor data points in $j$-th time window.  
Here, $n = p \cdot w$ and $d$ is the number of statistical features.

\begin{figure}[!t]
\centering
\includegraphics[scale=0.47]{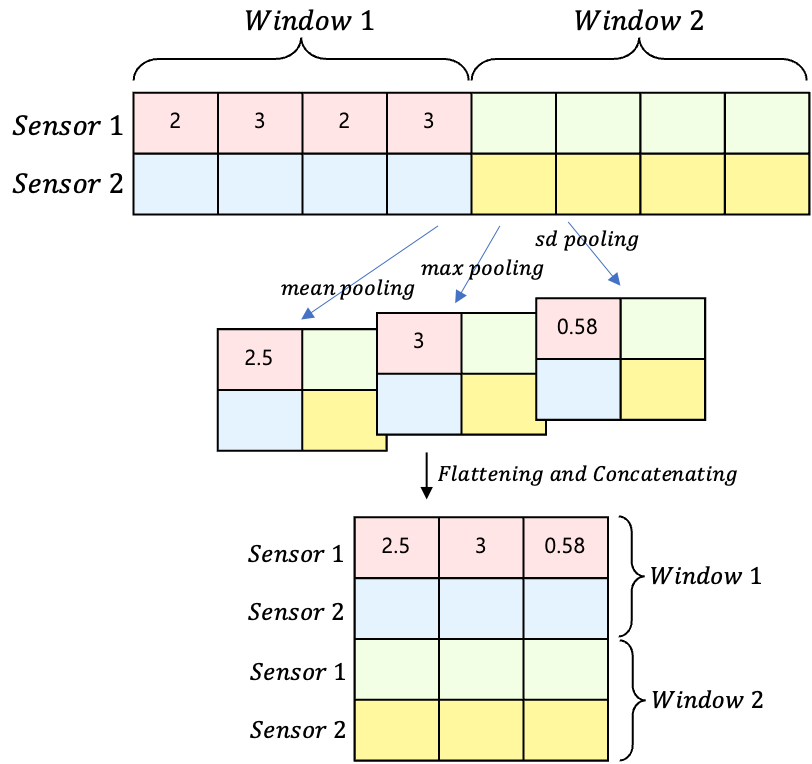}
\caption{Another interpretation of the process of statistical feature-based embedding.}
\label{fig:pooling}
\vskip -0.1in
\end{figure}

Another way to understand the process of the statistical feature embedding is by applying multiple statistical poolings together with flattening and concatenation, as illustrated in Fig \ref{fig:pooling}. Let $\mathbf Z \in \mathbb R^{p \times t}$ be the sensor data where its $(i,j)$-th element corresponds to the $i$-th sensor value at time $j$. We apply a statistical pooling layer on $\mathbf Z$, such as mean or maximum pooling, with a filter size of $(1, n_I)$ and a stride size of $(1, n_I)$. The pooling yields the matrix of dimension $(p,w)$, say $\mathbf Z_{\text{pool}} = [\mathbf z_1 , \ldots , \mathbf z_w]$. We then flatten $\mathbf Z_{\text{pool}}$ and obtain $\text{Flat}(\mathbf Z_{\text{pool}}) = [\mathbf z_1^T ~\ldots ~ \mathbf z_p^T]^T$. Repeating this process with $d$ different statistical poolings gives $d$ vectors: $\mathbf Z_{\text{pool}}^{[1]}, \ldots, \mathbf Z_{\text{pool}}^{[d]}$. Concatenating these vectors horizontally results in $\mathbf X_{\text{stat-emb}} = [\text{Flatten}(\mathbf Z_{\text{pool}}^{[1]}), \ldots, \text{Flatten}(\mathbf Z_{\text{pool}}^{[d]})]$, which is equivalent to \eqref{eq:x-stat-emb}.

%Then we can have $x_{ij}$ be the $k$-th statistical feature of $\{ s_{ij} \}, j\in I_j$, which is the statistical feature of $i$-th sensor data points in $j$-th time window. We organize the sensor's statistical features data so that each column represents each statistical feature and each row represents p-sensor k-time interval as follows:

\subsection{Window Positional Encoding}
The rows of $\mathbf X_{\text{stat-emb}}$ 
are sorted by the sequential order of time windows and then, for rows with the same time window, further sorted by sensors. The time windows are in the order of time, but within a time window, the rows do not contain any sequential information, which is distinct from $\mathbf X_{\text{emb}}$ in NLP. 

To inject the positional information respecting the nature of our $\mathbf X_{\text{stat-emb}}$, we propose the window positional encoding, giving the same position weight within the same window block. 
We define the window position encoding matrix $\mathbf P \in \mathbb R^{n \times p}$ as follows: 
\begin{equation*}
    \begin{aligned}
        & \mathbf P_{(pos, 2i)} = \sin ( \lceil pos/p\rceil /10000^{2i/d}) \\
        \text{and } & \mathbf P_{(pos, 2i+1)} = \cos ( \lceil pos/p \rceil /10000^{2i/d}),
    \end{aligned}
\end{equation*}
where $\lceil a \rceil$ is the nearest integer equal to or bigger than $a$. Similar to the original positional encoding introduced in the previous section, the matrix holds sequential information in time windows. In contrast to the original positional encoding, it provides consistent positional information within the same window block. Notice that $\lceil pos/p\rceil$ remains the same within the same time window. 

The positional information is injected in $\mathbf X_{\text{stat-emb}}$ by adding $\mathbf P$:
\begin{equation*} %\label{eq:X}
    \mathbf X = \mathbf X_\text{stat-emb} + \mathbf P.
\end{equation*}

\subsection{Transformer Block and Prediction}
We adopt an encoder transformer block from the Transformer. After then, we flatten the output matrix from the transformer block by the flattened layer. After that, a feed-forward neural network is used with a linear activation function for a regression problem such as building a VM model and a sigmoid activation function for a classification problem such as constructing an FD model.

\section{Experiments}\label{sec:real-analyses}

To validate the effectiveness of the proposed model, we conducted experiments with two real manufacturing datasets. We constructed two predictive models, the FD model and the VM model, on the two datasets, respectively. The two datasets consist of multivariate time-series sensor data, including temperature, humidity, and pressure, collected from the etching process of semiconductor manufacturing. The two datasets have different sets of sensor values. In semiconductor manufacturing, hundreds of unit processes are repeated. Therefore, even if the data is collected from the same process, the sensors used can vary. The first dataset includes classification labels (fault or normal). Thus, it is required to build a FD model. In contrast, the other dataset contains an measurement value. By predicting and monitoring this measurement value, engineers can determine the equipment's status and check the production quality. Therefore, for the second dataset, we build a VM model. While constructing predictive models using our proposed approach, we extracted six statistical feature from sensor data--mean, max, min, variance, range, and slope--for the statistical feature embedding.

\subsection{Baseline models}

There have been several approaches to handling sensor data. We compared our proposed method with representative methods in each of these approaches.
The first approach is extraction of statistical features from sensors using a window and constructed tabular data. Then, we fed this data into conventional machine learning models, including the least absolute shrinkage and selection operator (Lasso) \cite{tibshirani1996regression}, a random forest (RF) \cite{breiman2001random}, and extreme gradient boosting (Xgboost) \cite{chen2016xgboost}. 

The second approach involves deep learning models. To learn graphical features from time-series data, CNN, LSTM, and Bi-LSTM-based models have been predominantly utilized. For the CNN model, CNN for fault detection model (FDC-CNN), which was proposed for handling time-series data in manufacturing, was employed \cite{lee2017convolutional}. As for the LSTM and Bi-LSTM models, the sensor values for each time step were fed into each layer \cite{siami2019performance}. 

Finally, the following Transformer-based predictive models were employed. Sensor variables were processed as words in the encoder blocks (TF1) in the Transformer \cite{zhang2022deep}. In TF2 model, time information was treated as words in the encoder blocks \cite{peng2022fault}. Additionally, a neural network, utilizing the Transformer architecture and incorporating statistical features and LSTM layers, was employed (TF+LSTM) \cite{liu2020novel}.

\subsection{Case1: Fault detection problem}
\subsubsection{Dataset}
We utilized a real-world dataset obtained from a etching process of semiconductor manufacturer in South Korea. It contains 16 instances of faulty production and 8 instances of normal production. In addition, this dataset has 107 sensor variables. Our primary objective was to identify instances of faulty production using this dataset. The duration of the observations varies, with lengths ranging from 390 to 400 units. 

\subsubsection{Experimental design}
To conduct experiments properly, we split the total dataset into training and test data with an 80:20 ratio and stratified sampling. We applied both the proposed method and baseline models to the training data and evaluated their performance on the test data. Each experiment was repeated 10 times, and we calculated the average of performance measures across 10 repetitions. We selected three performance measures: F1-score (F1), accuracy (Acc), and the area under the curve (AUC). Finally, we set the window size to 20 to extract statistical features for the statistical feature embedding in our proposed model.

%The F1-score represents the harmonic mean of recall and precision and AUC measures the area under the receiver operating characteristic curve. 

\subsubsection{Result}
Table \ref{table1} presents the comparison results between the baseline models and the proposed model on the FD problem. Due to the limited dataset size, the ML-based models outperformed the DL-based models. This outcome reaffirmed that achieving high performance in deep learning requires a substantial amount of data \cite{munir2018deepant,krishna2019deep,he2020model}. Furthermore, as previously suggested in other studies, the TF-based models demonstrated superior performance over the DL-based models. Notably, the proposed model outperformed all baseline models in terms of Acc and F1. The effectiveness of statistical features in a manufacturing sensor dataset was confirmed through the results of both ML-based and proposed models. 

\begin{figure*}[t!]
\centerline{\includegraphics[scale=0.28]{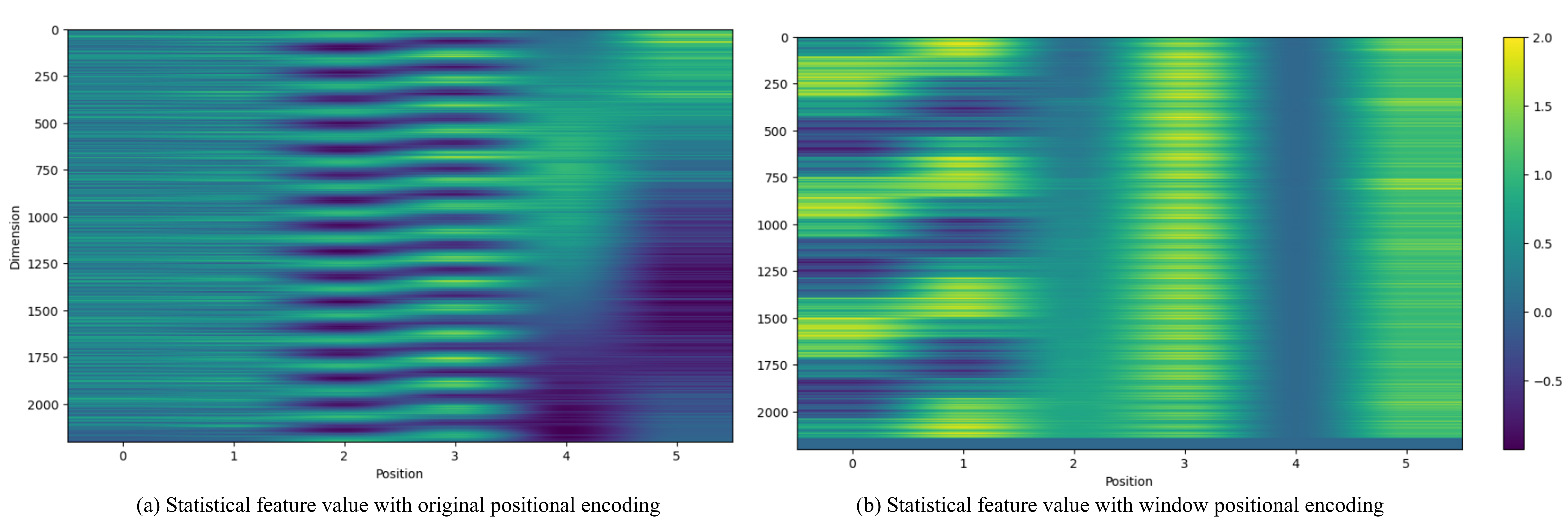}}
\caption{Visualization of statistical feature embedding with original positional encoding and window positional encoding.}
\label{fig8}
\vskip -0.1in
\end{figure*}

Table \ref{table2} illustrates the performance of the proposed model with window positional encoding (WPE) and original positional encoding (PE), varying the number of encoder layers and multi-head attention (MHA). The proposed model with WPE exhibited superior performance compared to that with PE. WPE allows the model to capture temporal information within the statistical feature embedding. Fig \ref{fig8} shows the visualization of statistical feature embedding with PE and WPE. In the first experiment, the number of window groups is 20; thus, it was observed that the embedding matrix was also divided into 20 groups, as shown in Fig \ref{fig8}.(b). Interestingly, the proposed model with PE also outperformed other baseline models. This is attributed to PE providing approximate temporal information on the embedding matrix, even though it does not precisely indicate the positional information. 
Additionally, there were no significant differences as we increased the number of encoder layers and MHA.

\begin{table}[t]
\caption{Performance comparison of the baseline models and proposed model in the fault detection problem.}
\label{table1}
\setlength{\tabcolsep}{8 pt} % m
\begin{center}
\begin{small}
\vskip -0.1in
%  \begin{sc}
\begin{tabular}{l l c c c }
\toprule
  \multicolumn{1}{c}{} & \multicolumn{1}{c}{Model} & \multicolumn{1}{c}{Acc} & \multicolumn{1}{c}{F1} & \multicolumn{1}{c}{  AUC   }  \\
\midrule

\multicolumn{1}{c}{\multirow{4}{*}{ML-based}} & \multicolumn{1}{c}{\multirow{1}{*}{LASSO }}& 0.775 & 0.825 & $\mathbf{0.916}$ \\

 \cmidrule{2-5} 
& \multicolumn{1}{c}{\multirow{1}{*}{Xgboost}} & 0.785& 0.827  & 0.860 \\

 \cmidrule{2-5} 
& \multicolumn{1}{c}{\multirow{1}{*}{RF}} & 0.741 & 0.752  & 0.852  \\
\midrule

\multicolumn{1}{c}{\multirow{4}{*}{DL-based }} & \multicolumn{1}{c}{\multirow{1}{*}{FDC CNN }}& 0.655 & 0.715 &  0.720 \\

 \cmidrule{2-5} 
& \multicolumn{1}{c}{\multirow{1}{*}{LSTM}} & 0.665 & 0.706   & 0.655   \\

 \cmidrule{2-5} 
& \multicolumn{1}{c}{\multirow{1}{*}{Bi-LSTM}} & 0.645  & 0.725   & 0.676 \\

\midrule

\multicolumn{1}{c}{\multirow{5}{*}{TF-based }} & \multicolumn{1}{c}{\multirow{1}{*}{TF1}}& 0.755 & 0.775 & 0.836\\

 \cmidrule{2-5} 
& \multicolumn{1}{c}{\multirow{1}{*}{TF2}} & 0.746 & 0.760  & 0.829 \\

 \cmidrule{2-5}
& \multicolumn{1}{c}{\multirow{1}{*}{TF+LSTM}} & 0.728 & 0.732  & 0.843 \\
 \cmidrule{2-5}
 
& \multicolumn{1}{c}{\multirow{1}{*}{Ours}} & $\mathbf{0.805}$ & $\mathbf{0.857}$  & 0.900 \\

\bottomrule
\end{tabular}
%  \end{sc}
\end{small}
\end{center}
\vskip -0.2in
\end{table}

\begin{table}[t]
\caption{Performance comparison of the proposed model with window positional encoding and original positional encoding in the fault detection problem.}
\label{table2}
\setlength{\tabcolsep}{2.5pt} % m
\begin{center}
\begin{small}
\vskip -0.1in
%  \begin{sc}
\begin{tabular}{l l c c c c c c }
\toprule
  \multicolumn{1}{c}{} & \multicolumn{3}{c}{WPE} & \multicolumn{3}{c}{PE}  \\
\midrule

  \multicolumn{1}{c}{} & \multicolumn{1}{c}{Acc} & \multicolumn{1}{c}{F1} & \multicolumn{1}{c}{AUC} & \multicolumn{1}{c}{Acc} & \multicolumn{1}{c}{F1} & \multicolumn{1}{c}{AUC}\\
\midrule
  
  \multicolumn{1}{c}{ \# of layer=1, \# of MHS=1} & 0.805 & 0.857 & 0.900 & 0.775 & 0.834 & 0.876\\
\midrule
  \multicolumn{1}{c}{ \# of layer=4, \# of MHS=1} & 0.790 & 0.845 & 0.890 & 0.780 & 0.830 & 0.870\\
\midrule
  \multicolumn{1}{c}{ \# of layer=4, \# of MHS=4} & 0.795 & 0.855 & 0.895 & 0.770 & 0.825 & 0.860\\

\bottomrule
\end{tabular}
%  \end{sc}
\end{small}
\end{center}
\vskip -0.2in
\end{table}

\subsection{Case2: Virtual metrology problem}
 \subsubsection{Dataset}
 In the virtual metrology problem, we employed an actual semiconductor dataset gathered from South Korea over a period of 15 days. It contains 488 observations and 51 sensor variables, and the target variable is the electrical measurement value. The sensor variables are measured at 45-47 time points for each day.
 
\subsubsection{Experimental design}
To assess the performance of the VM model, we employed a window sliding validation approach. Initially, we use the data from the first 5 days for training and the data from the following day for testing. Subsequently, the window shifts by 1 day, and this process continues until a total of 10 days' worth of data has been used for testing. For the performance measures, we used mean squared error (MSE) and mean absolute percentage error (MAPE), and we reported the average performance value for all test days. Finally, we set the window size to 10 to extract statistical features.

\subsubsection{Result}
\begin{figure*}[t!]
\centerline{\includegraphics[scale=0.54]{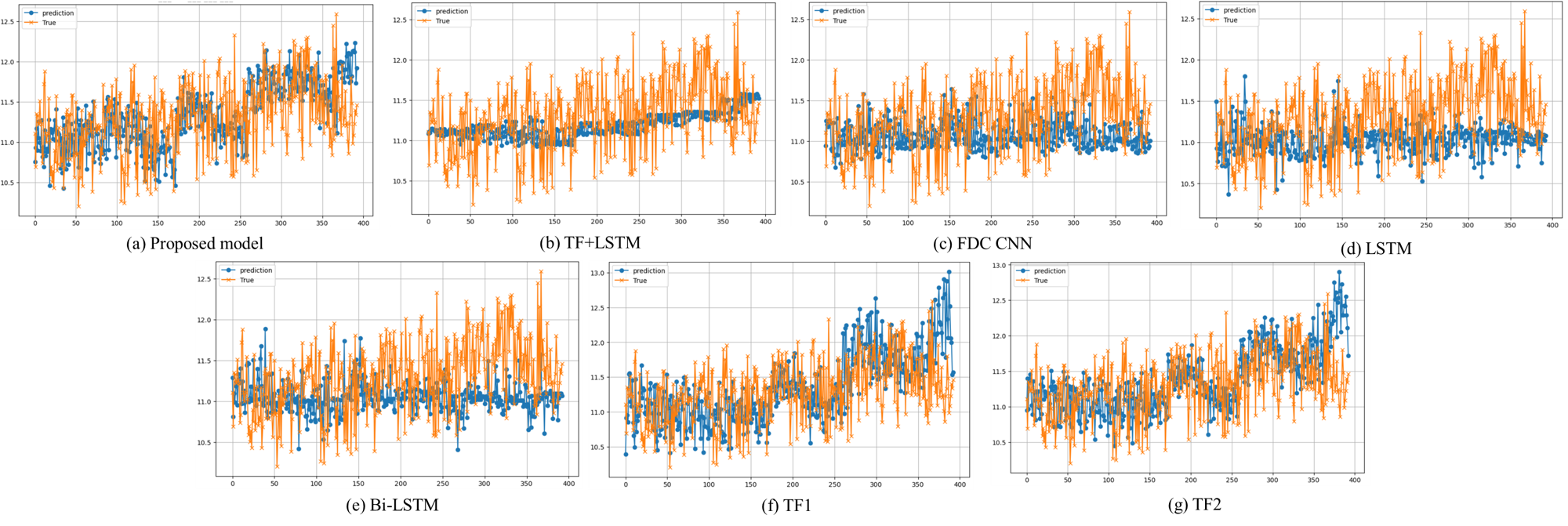}}
\caption{Plot of prediction values and actual values for all the models.}
\label{fig7}
\vskip -0.2in
\end{figure*}

Table \ref{table3} presents the comparison results between the baseline models and the proposed model for the VM problem. The proposed model demonstrated excellent performance compared to the baseline models except the TF-LSTM. The TF-LSTM model showed similar results to our model. However, as shown in Fig \ref{fig7}.(b), the TF-LSTM model predicted values as the mean of the training target values. In contrast, the prediction values of the proposed model closely followed the trend of actual values, as shown in Fig \ref{fig7}.(a). The proposed model has a small capacity (with 39,065 parameters), while the TF-LSTM has a large capacity (with 6,782,689 parameters). The TF-LSTM model exhibited an overfitting problem due to its excessive capacity compared to the dataset size. Therefore, in this context, although the performance measures appear similar, it is more reasonable to use our model. In fact, this phenomenon was observed in all DL-based models with large capacity, as shown in Fig \ref{fig7}.(b) to (e). The TF1 and TF2 models could mitigate the overfitting problem. However, their MSE and MAPE were significantly higher than those of our model. Furthermore, similar to the first experiment, the proposed model with WPE outperformed the model with PE. There were no significant differences as we increased the number of encoder layers or/and MHA.

\begin{table}[t]
\caption{Performance comparison of the baseline models and proposed model in the virtual metrology problem.}
\label{table3}
\setlength{\tabcolsep}{11 pt} % m
\begin{center}
\begin{small}
\vskip -0.1in
%  \begin{sc}
\begin{tabular}{l l c  c }
\toprule
  \multicolumn{1}{c}{} & \multicolumn{1}{c}{Model} & \multicolumn{1}{c}{MSE} & \multicolumn{1}{c}{MAPE}  \\
\midrule

\multicolumn{1}{c}{\multirow{4}{*}{ML-based}} & \multicolumn{1}{c}{\multirow{1}{*}{LASSO }}& 0.451 & 0.076  \\

 \cmidrule{2-4} 
& \multicolumn{1}{c}{\multirow{1}{*}{Xgboost}} & 0.255& 0.042   \\

 \cmidrule{2-4} 
& \multicolumn{1}{c}{\multirow{1}{*}{RF}} & 0.231 & 0.034    \\
\midrule

\multicolumn{1}{c}{\multirow{4}{*}{DL-based }} & \multicolumn{1}{c}{\multirow{1}{*}{FDC CNN }}& 0.301 & 0.039  \\

 \cmidrule{2-4} 
& \multicolumn{1}{c}{\multirow{1}{*}{LSTM}} & 0.299 & 0.039      \\

 \cmidrule{2-4} 
& \multicolumn{1}{c}{\multirow{1}{*}{Bi-LSTM}} & 0.289  & 0.038    \\

\midrule

\multicolumn{1}{c}{\multirow{5}{*}{TF-based }} & \multicolumn{1}{c}{\multirow{1}{*}{TF1}}& 0.310 & 0.039 \\

 \cmidrule{2-4} 
& \multicolumn{1}{c}{\multirow{1}{*}{TF2}} & 0.283 & 0.037   \\

 \cmidrule{2-4}
& \multicolumn{1}{c}{\multirow{1}{*}{TF+LSTM}} & $\mathbf{0.215 }$& $\mathbf{0.033}$   \\

 \cmidrule{2-4}
 
& \multicolumn{1}{c}{\multirow{1}{*}{Ours}} & $\mathbf{0.223}$& $\mathbf{0.034}$   \\

\bottomrule
\end{tabular}
%  \end{sc}
\end{small}
\end{center}
\vskip -0.2in
\end{table}

\begin{table}[t]
\caption{Performance comparison of the proposed model with window positional encoding and original positional encoding in the virtual metrology problem.}
\label{table5}
\setlength{\tabcolsep}{5 pt} % m
\begin{center}
\begin{small}
\vskip -0.1in
%  \begin{sc}
\begin{tabular}{l l c c c c }
\toprule
  \multicolumn{1}{c}{} & \multicolumn{2}{c}{WPE} & \multicolumn{2}{c}{PE}  \\
\midrule

  \multicolumn{1}{c}{} & \multicolumn{1}{c}{MSE} & \multicolumn{1}{c}{MAPE} & \multicolumn{1}{c}{MSE} & \multicolumn{1}{c}{MAPE}\\
\midrule
  
  \multicolumn{1}{c}{ \# of layer=1, \# of MHS=1} & 0.223 & 0.033 & 0.251 & 0.036 \\
\midrule
  \multicolumn{1}{c}{ \# of layer=4, \# of MHS=1} & 0.233 & 0.035 & 0.247 & 0.035 \\
\midrule
  \multicolumn{1}{c}{ \# of layer=4, \# of MHS=4} & 0.219 & 0.034 & 0.260 & 0.367 \\

\bottomrule
\end{tabular}
%  \end{sc}
\end{small}
\end{center}
\vskip -0.2in
\end{table}

\subsection{Performance comparison based on model capacity}
\begin{figure}[!t]
\centering
\includegraphics[scale=0.27]{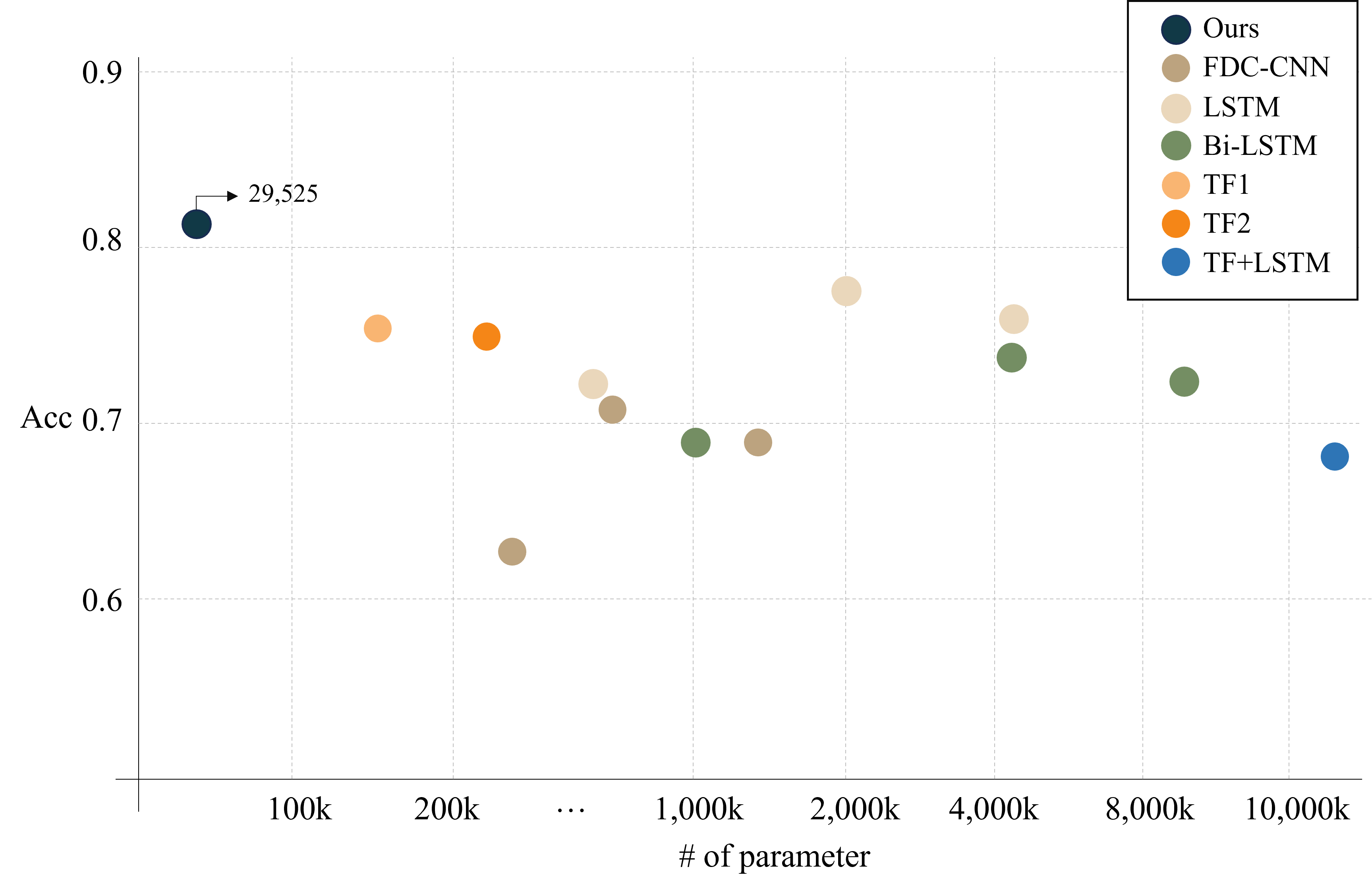}
\caption{Plot of performance of all the models against the number of parameter for the FD problem. }
\label{fig5}
\vskip -0.1in
\end{figure}

\begin{figure}[!t]
\centering
\includegraphics[scale=0.27]{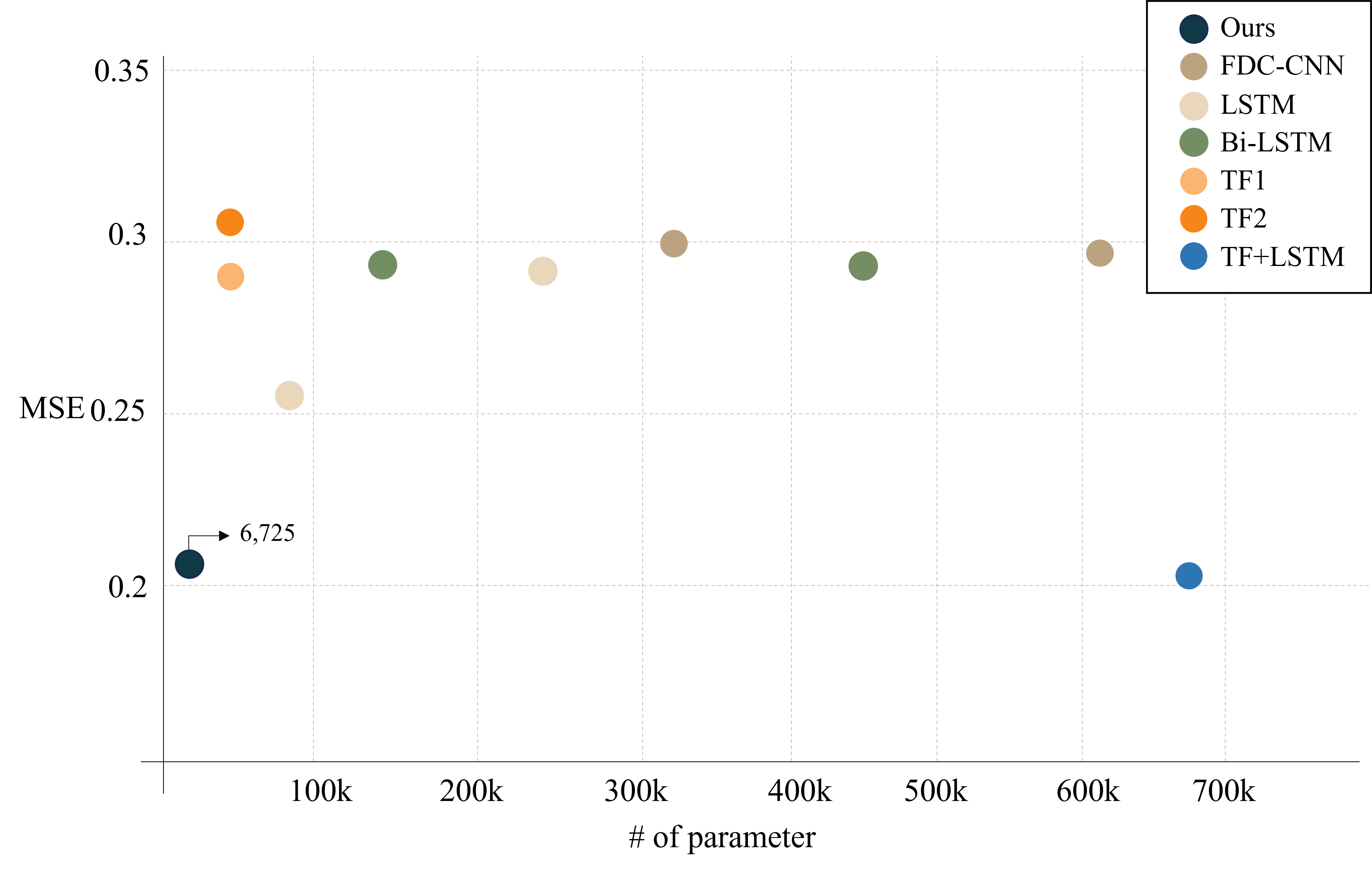}
\caption{Plot of performance of all the models against the number of parameter for the VM problem.}
\label{fig6}
\vskip -0.1in
\end{figure}

Fig.\ref{fig5} and Fig.\ref{fig6} plot the performance of all the models against the number of parameters for the FD problem and VM problem, respectively. For a more accurate comparison, we conducted experiments by adjusting the number of layers and filters in the DL-based models. In other words, we compared both deeper models and shallower models. We can confirm that as the number of parameters increased, the model's performance became poorer. In particular, the FD model with the smallest capacity showed the best performance. Similarly, the proposed VM models exhibited excellent performance with only 6,726 parameters. It is noteworthy that the number of parameters to be trained in the proposed model is at least 1/10 to 1/100 compared to other DL-based models. 

\section{Discussion}
This paper introduce a novel Transformer for the manufacturing sensor domain by incorporation of the statistical feature embedding and WPE. Our real data analyses support the effectiveness of the proposed model.

The improved performance of our proposed model could be attributed to the following reason. Thanks to the statistical feature embedding, our proposed Transformer is highly efficient in terms of parameter usage, requiring significantly fewer parameters than other deep learning architectures, as shown in Fig.\ref{fig5} and Fig.\ref{fig6}. The number of trainable parameters in the Transformer depends on the number of word dimensions. In other Transformer-based models, the number of word dimensions is the same as the time length or the number of sensors, both of which are usually large. Therefore, when the time length is long or the number of sensor variables is a lot, they require a large number of trainable parameters. In our study, the number of word dimensions is equal to the number of statistical features, which is smaller than the number of sensors and time points. As a result, our model has a smaller capacity and exhibit an excellent performance when the data sample size is small. This efficiency is particularly advantageous in the context of the manufacturing sensor domain, where data samples are naturally limited due to frequent changes in manufacturing processes.

%\st{Secondly, the statistical feature embedding empowers the Transformer to effectively learn both sensor and time-series information. Each row in the statistically embedded matrix represents a combination of sensor and time variables, facilitating the model to learn from both variables. Furthermore, the statistical feature embedding effectively condenses sensor information into a smaller number of time points by considering time windows and statistical features. This process reduces unnecessary variability, resulting in enhancing the model's performance.}

 %A common approach when applying the Transformer to time-series sensor data involves the utilization of all sensor values at every time point. For example, 
%organizing the sensor data itself into a matrix form. 
%each column corresponds to a sensor variable (or a time variable) and each row corresponds to a time variable (a sensor variable). In contrast, our approach differs in that we calculate statistical features for each sensor over a time window and then organize these features into a matrix from, as described in \eqref{eq:x-stat-emb}. 

%However, unlike previous approaches, our proposed model architecture is based on the Transformer. To the best of our knowledge, this is the first paper to employ statistical features in deep learning or Transformer model architectures.

\begin{figure}[!t]
\centering
\includegraphics[scale=0.25]{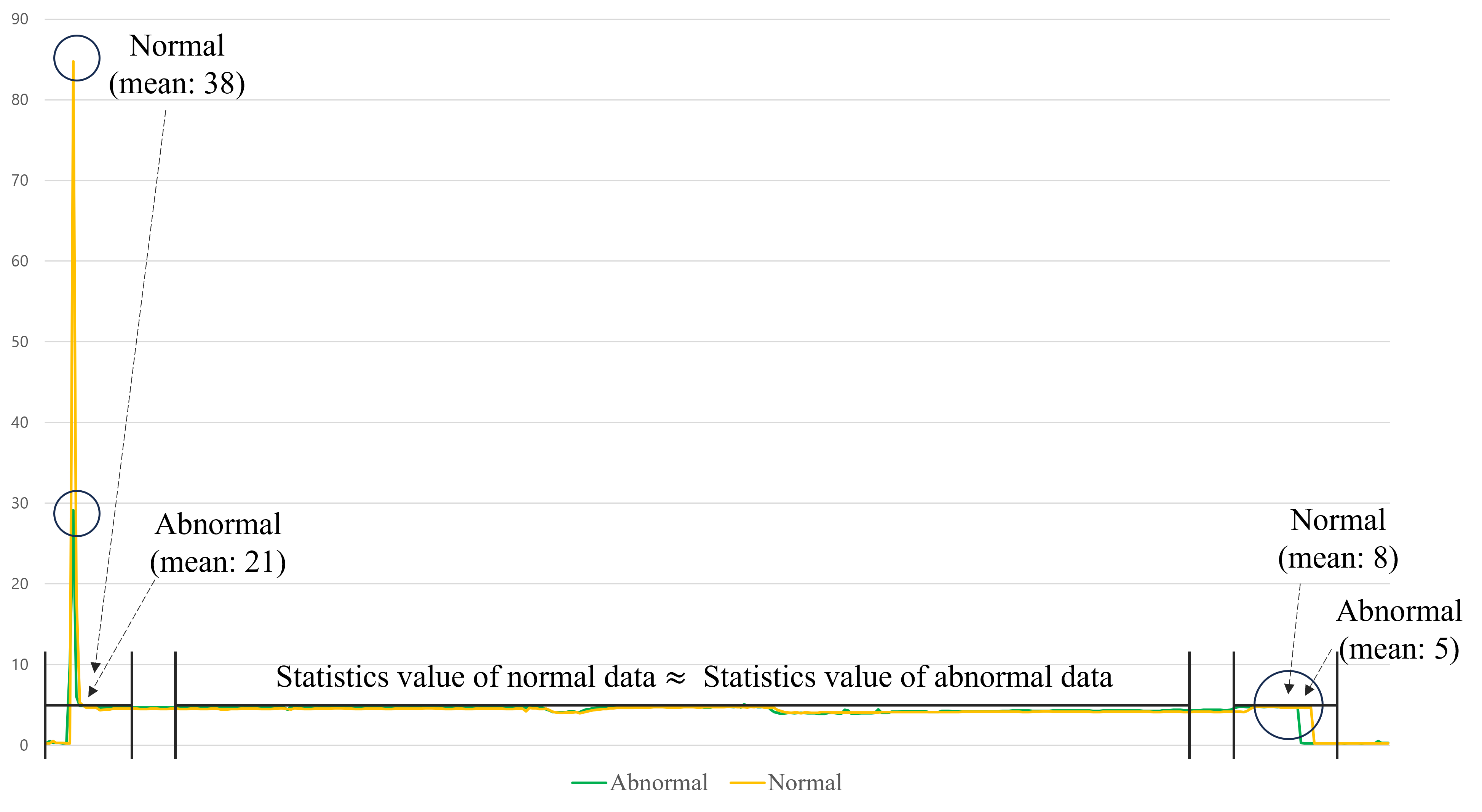}
\caption{Specific sensor values of normal and abnormal data}
\label{fig:why_stat-embed}
\vskip -0.1in
\end{figure}

%Before delving into the rationale behind this new approach, it's important to understand the characteristics of time-series sensor data in general. 
The use of statistical features has been adopted in machine learning-based approaches, where these features are extracted and employed as inputs in the models. Our approach aligns with this approach as we utilize statistical features. To comprehend why the use of statistical features shows good performance, it is essential to understand the characteristics of time-series sensor data.
Fig \ref{fig:why_stat-embed} illustrates specific sensor values of normal and abnormal production. The x-axis represents time and the y-axis represents sensor values. We can see that the discernible differences between normal and abnormal productions are confined to few particular time points. Both normal and abnormal productions reach their maximum values simultaneously at an early time point, with the normal observation exhibiting a higher maximum value compared to the abnormal one. Towards the end, the sensor values for the normal sample drop to near-zero slightly earlier than the abnormal one. Excluding these specific time points, distinguishing between normal and abnormal productions is challenging. That is, the mean values at the first time windows are sufficient to distinguish between these two data categories. Despite the large number of time points at which sensor data is measured, most of the time points are irrelevant for distinguishing between normal and abnormal states or predicting their respective output values. This suggests that predictive models for sensor data do not require all data points, and aggregating sensor data points within a time window using statistics such as mean, max, and variance is sufficient. Our real data analyses' results in Section \ref{sec:real-analyses} support the effectiveness of our proposed statistical feature embedding by showing good performance compared to the existing approach.

%a complicated and complex model.  
%Our approach is aligned with this approach. We aggregate sensor data points within a time window using statistics such as mean, max, and standard deviation. This aggregation helps reduce the extraneous variation caused by the numerous time points by reducing the number of time points $t$ to the number of time windows $w$. 
%While some sensor information can be lost due to the nature of the aggregation, statistical features still capture the underlying differences between normal and abnormal productions. For instance, in the example of Fig \ref{fig:why_stat-embed}, the mean values at the first and last time windows are sufficient to distinguish between these two data categories. 
%Our real data analyses' results in Section \ref{sec:real-analyses} support the effectiveness of our proposed statistical feature embedding by showing good performance compared to the existing approach.

In our proposed Transformer, the choice of which statistical features to use should be made by the user prior to training the model, similar to determining hyper parameter values. If this process could be automatically learned by the model, it would be advantageous in improving the model performance further more. One way to view this research question is as follows. Choosing statistical features is likened to selecting statistical pooling layers, as we discussed in Subsection \ref{subsec:stat-embed}. Consequently, this question can be reformulated as the search for the optimal pooling layers. Exploring this question in the direction of selecting optimal pooling layers would be an intriguing avenue for future research.

\section{Conclusion}

With the simultaneous development of manufacturing process technology and artificial intelligence, a gap exists between these two fields. Various deep learning models have been increasingly developed to be more complex for real-world applications. Consequently, they require vast datasets to demonstrate their performance. However, in the manufacturing industry, where processes rapidly change, it can be challenging to collect a significant amount of data. Additionally, when the process changes, they cannot use machine learning or deep learning models to detect defects or manage various measurement values until a sufficient amount of data is collected.

To address this issue, our study proposes a novel predictive model based on the Transformer architecture, incorporating statistical feature embedding and WPE. Our study is motivated by the following observations: First, statistical features are highly effective for predicting the target variable in the manufacturing sensor domain. Second, the Transformer architecture can simultaneously learn both sequential information and sensor data using the proposed statistical feature embedding. Lastly, with the statistical feature embedding, we can significantly reduce the number of parameters to be trained in the model.

The proposed predictive model can sequentially learn statistical feature information and has exhibited excellent performance in two types of problems with the smallest capacity. The experimental results with actual datasets support the effectiveness of the proposed model. Our model can be applied in various manufacturing domains where the amount of available data is limited. Moreover, since our model is based on neural networks, it can be adapted for online learning. As more data are collected, the model can be updated, and performance can be improved using online learning.

In future research, it is necessary to identify the optimal statistical features. As previously discussed, the optimal statistical features may vary depending on the industry domain. By optimizing the multiple pooling layers, we expect to effectively build the predictive model.

%\section*{Acknowledgments}
%This work was partly carried out during a visit to DIMACS partially enabled by NSF award CCF-144575.

%\section*{Acknowledgments}
%This work was partly carried out during a visit to DIMACS partially enabled by NSF award CCF-144575.

\bibliographystyle{IEEEtran}
\bibliography{mybibfile}

\end{document}